\documentclass[11pt,a4paper]{article}

\usepackage[a4paper,top=2.0cm,bottom=2.0cm,left=2.05cm,right=2.05cm]{geometry}
\usepackage[T1]{fontenc}
\usepackage[utf8]{inputenc}
\usepackage{lmodern}
\usepackage{amsmath,amssymb,bm}
\usepackage{booktabs,longtable,tabularx,array,multirow,makecell}
\usepackage[table]{xcolor}
\usepackage{graphicx}
\usepackage{enumitem}
\usepackage{hyperref}
\usepackage{url}
\usepackage{caption}
\usepackage{float}
\usepackage{setspace}
\usepackage{titlesec}
\usepackage{fancyhdr}
\usepackage{tikz}
\usetikzlibrary{arrows.meta,positioning,fit,calc,shapes.geometric}

\definecolor{softblue}{RGB}{232,240,250}
\definecolor{softorange}{RGB}{252,239,222}
\definecolor{softgreen}{RGB}{231,245,235}
\definecolor{softgray}{RGB}{245,245,245}
\definecolor{deepblue}{RGB}{36,86,145}
\definecolor{deeporange}{RGB}{190,105,34}

\hypersetup{
  colorlinks=true,
  linkcolor=deepblue,
  citecolor=deepblue,
  urlcolor=deepblue,
  pdfauthor={Jiawei Liu, Jiacheng Guo, Tian Zhang, Yiwei Xu, Juan Wang, Jinlin Fan, Bowen Xiao},
  pdftitle={Security of Foundation-Model-Powered Embodied Agents: Attack Surfaces, Attacks, Defenses, and Evaluation}
}

\setlist[itemize]{leftmargin=2em,itemsep=0.25em,topsep=0.3em}
\setlist[enumerate]{leftmargin=2.2em,itemsep=0.25em,topsep=0.3em}
\titleformat{\section}{\Large\bfseries}{\thesection}{0.7em}{}
\titleformat{\subsection}{\large\bfseries}{\thesubsection}{0.7em}{}
\titleformat{\subsubsection}{\normalsize\bfseries}{\thesubsubsection}{0.6em}{}

\newcommand{\surveyRQ}[1]{\textbf{RQ#1}}

\title{\textbf{Security of Foundation-Model-Powered Embodied Agents:\\Attack Surfaces, Attacks, Defenses, and Evaluation}}
\author{
{\rm Jiawei Liu}\\
Wuhan University
\and
{\rm Jiacheng Guo}\\
Wuhan University
\and
{\rm Tian Zhang}\\
Wuhan University
\and
{\rm Yiwei Xu}\\
Wuhan University
\and
{\rm Juan Wang}\\
Wuhan University
\and
{\rm Jinlin Fan}\\
Wuhan University
\and
{\rm Bowen Xiao}\\
Wuhan University
} 
\date{}

\begin{document}
\maketitle

\begin{abstract}
Foundation models are increasingly used for perception, reasoning, planning, and action generation in embodied agents, creating security risks that can propagate from digital inputs to physical behavior. Existing surveys often organize threats by mechanisms such as jailbreaks, prompt injection, backdoors, poisoning, or adversarial examples, but these categories do not consistently identify where an adversary first enters the embodied control loop. We present a trust-boundary-centric survey of foundation-model-powered embodied-agent security. Using a first-compromised-trust-boundary principle, we separate attack surface from attack mechanism and organize the system into five layers and twelve attack surfaces spanning the model supply chain, user instructions, context and memory, physical semantic environments, multimodal perception, world state, internal reasoning, task planning, action interfaces, middleware, multi-agent communication, and execution control. Based on 58 attack records and 61 defense records collected through August 15, 2026, we analyze representative attacks, cross-layer propagation, defense placement, and evaluation practices. Our quantitative analysis shows that attack research is concentrated on multimodal perception and action interfaces, while defenses are especially concentrated on action-level and runtime protection. Context and long-term memory, middleware and networking, world-state integrity, and multi-agent trust remain comparatively underexplored. We conclude with open challenges in state provenance, compositional defenses, long-horizon attack propagation, physical realizability, Byzantine multi-robot behavior, and unified closed-loop evaluation.
\end{abstract}


\tableofcontents
\newpage

\section{Introduction}

\subsection{Foundation models are restructuring the robotic control loop}
The rapid development of LLMs, VLMs, and VLAs is transforming robotics from a relatively closed architecture built around predefined task interfaces, specialized perception modules, and hand-engineered planners into an open architecture that accepts free-form language, multimodal observations, external knowledge, and learned action policies. LLMs increasingly perform task decomposition, code generation, tool invocation, and high-level planning; VLMs jointly map visual observations and language goals into semantic representations; and VLAs further compress the conventional perception--planning--control stack by generating action tokens, continuous controls, or action chunks directly from visual and linguistic conditions. World models and world-action models introduce yet another layer by predicting future states and action consequences before control decisions are issued. Recent surveys on foundation-model-powered robots, LLM-agent security, VLA safety, and world-model security describe this structural transition from complementary perspectives\cite{S01,S02,S03,S04}.

The same capabilities that make foundation models useful also expand the set of inputs that a robot may treat as decision-relevant. Conventional robots typically consume structured goals, sensor streams, and fixed APIs. Foundation-model-powered robots may additionally read natural-language instructions, scene text, retrieved web or RAG content, interaction history, long-term memory, messages from other agents, and heterogeneous multimodal observations. Once such information becomes part of the control loop, data that would ordinarily be considered ``semantic content'' can acquire physical authority. Consequently, an adversary may not need to compromise a motor controller, ROS node, or actuator directly; manipulating information trusted by the model can be sufficient to induce the system to generate an adversary-desired behavior on its own.

\subsection{Why embodied-agent security differs from text-only LLM security}
A harmful output from a text-only LLM typically remains in the information domain. In an embodied system, model outputs can be translated through code, APIs, skill libraries, motion planners, and controllers into physical actions. The same jailbreak that produces unsafe text in a chatbot may therefore become target substitution, navigation deviation, hazardous manipulation, trajectory drift, action freezing, or denial of service when the model is embedded in a robot. BadRobot highlighted the mismatch between linguistic refusal behavior and action safety\cite{A004}, while RoboPAIR automated adversarial prompt search with the explicit objective of inducing unsafe robot behavior under different model-access assumptions\cite{A007}. These findings show that embodied attacks cannot be evaluated solely by whether a model ``complies'' with a malicious request; security analysis must trace the effect through planning, action generation, and final environment state.

Embodied systems also introduce risks that are absent or much less prominent in text agents. First, environmental and sensor attacks must satisfy physical realizability constraints involving view angle, distance, occlusion, lighting, and dynamics. Second, continuous control creates temporal accumulation: a small error at one step may alter the next observation and amplify over a long horizon. Third, mistakes can have irreversible physical consequences such as collision, dropping objects, or unsafe contact. Fourth, runtime defenses operate under strict latency budgets. Fifth, heterogeneous representation changes---from pixels to semantic state, from language to plans, and from plans to executable actions---create multiple privilege transitions. Embodied AI security is therefore better viewed as \emph{cyber--physical agent security} rather than a simple union of LLM jailbreaks, adversarial vision, and traditional robot cybersecurity.

\subsection{What existing surveys cover, and what remains missing}
The SoK on security and privacy of foundation-model-powered robots organizes risks using Foundation Model, Embodied System, Supporting Ecosystem, and Governance (F--E--S--G) boundaries and encodes studies by target, lifecycle stage, mechanism, access, and effect\cite{S01}. The survey \emph{Toward Secure LLM Agents} emphasizes information flow, delegated authority, and persistent state across the agent lifecycle\cite{S02}. The VLA safety survey separates training-time and inference-time attacks and defenses while highlighting supply-chain poisoning, multimodal attacks, long-horizon error propagation, and real-time constraints\cite{S03}. The world-model security survey further introduces state grounding, learned dynamics, trajectory evaluation, execution feedback, and adaptation as security-relevant lifecycle stages\cite{S04}. These works provide important methodological foundations for the present study.

However, when the goal is to answer ``where, exactly, is a particular embodied attack entering the system?'', classifications based primarily on mechanisms or model types can remain ambiguous. A backdoor may be implanted during training but triggered by language, a physical object, a state pattern, or an action chunk. Prompt injection may come from the current user, from a screen in the environment, from a RAG document, or from another robot. An adversarial patch may primarily corrupt perception, but it may also be optimized to hijack chain-of-thought or continuous action generation. If labels such as ``backdoor'', ``prompt injection'', or ``patch'' are treated as top-level attack surfaces, the distinction between \emph{how an attack is implemented} and \emph{where it first crosses a trust boundary} is lost.

\subsection{Our perspective, research questions, and contributions}
We take a systems-security perspective centered on trust boundaries. We define an \textbf{attack surface} as the earliest boundary at which an adversary can directly act on a system representation and cause that representation to become untrustworthy. An \textbf{attack mechanism} is the technical method used to achieve the compromise. A \textbf{propagation path} captures subsequent changes in representation and control flow, while an \textbf{effect surface} denotes downstream components that are affected but are not the adversary's first direct point of control. This formulation follows a \emph{first compromised trust boundary} principle and permits the same mechanism to appear at multiple surfaces while allowing the same surface to host multiple mechanisms.

We study five research questions:
\begin{itemize}
  \item \surveyRQ{1}: How should a foundation-model-powered embodied agent be represented as a unified closed-loop system, and where should its trust boundaries be drawn?
  \item \surveyRQ{2}: Which trust boundaries are most frequently targeted by current attacks, and how do attack mechanisms propagate across layers into physical execution?
  \item \surveyRQ{3}: Where are existing defenses deployed, and does defense coverage match the distribution of exposed attack surfaces?
  \item \surveyRQ{4}: Which metrics are currently used at the model, planning, action, and physical layers, and what closed-loop metrics are still missing?
  \item \surveyRQ{5}: As long-term memory, world models, end-to-end VLAs, multi-robot collaboration, and cloud-edge integration become common, which security problems should receive priority?
\end{itemize}

Our contributions are fourfold. First, we model embodied agents as a closed information and control loop extending from development and supply chain to physical execution, explicitly distinguishing attack entry, propagation, and consequence. Second, we introduce a five-layer, twelve-surface taxonomy and decision rules that make mechanisms such as jailbreak, backdoor, poisoning, and sensor spoofing orthogonal to system entry points. Third, we provide a quantitative landscape analysis based on 58 attack records and 61 defense records, exposing asymmetries between attack and defense attention. Fourth, we connect taxonomy to evaluation and deployment, and derive a research agenda around closed-loop attack success, long-horizon causal propagation, physical realizability, provenance, and compositional defense.

\section{Survey Scope and Literature Methodology}

\subsection{Scope and study population}
We focus on embodied systems in which an LLM, VLM, VLA, world model, or world-action model participates materially in perception, state construction, task planning, policy generation, tool use, or action control. The scope includes hierarchical LLM-planner-plus-skill-executor architectures, VLM-based navigation, end-to-end VLA manipulation, LLM-integrated mobile robots, multi-robot LLM coordination, and the ROS/ROS2, networking, and cloud-model infrastructure that directly couples to these systems.

To avoid indiscriminately including all conventional robot-security work, we use three inclusion tiers. \textbf{A--Core} studies directly investigate attacks or defenses involving LLM/VLM/VLA/WAM components in an embodied control loop. \textbf{B--Adjacent} studies primarily address VLA, autonomous driving, vision--language navigation, or mechanisms that are directly transferable to the target systems. \textbf{C--System Extension} studies may not center on a foundation model but directly cover ROS, networking, multi-robot, or supply-chain boundaries that form part of the deployed system. This design keeps the survey focused while preserving infrastructure risks that would otherwise be artificially separated from foundation-model-powered robotics.

\subsection{Corpus and temporal boundary}
The corpus used in this version contains 58 attack records and 61 defense records through August 15, 2026. The attack corpus contains 50 A--Core, 3 B--Adjacent, and 5 C--System Extension records; the defense corpus contains 36 A--Core, 22 B--Adjacent, and 3 C--System Extension records. We use the foundation-model-powered robot SoK and VLA safety survey as major seed sources, cross-check them against LLM-agent security, LLM-controlled robotics risk surveys, and the world-model security survey, and include newer work from mid-2026\cite{S01,S02,S03,S04,S05}.

The term ``near-exhaustive'' should not be interpreted as a literal guarantee of every paper in a rapidly changing area. Many 2026 studies are first released as arXiv preprints, and venue assignments or experimental details may change. We therefore treat the corpus as a dynamic snapshot rather than a permanently closed bibliography.

\subsection{Coding fields and the multi-label principle}
Each attack record is coded with at least year, publication status, model type, inclusion tier, lifecycle stage, attack mechanism, attack-surface code, primary target module, attacker access, security objective, physical/real-world validation, evaluation platform, method summary, and main result. Defense records additionally capture defense stage, defense mechanism, covered surfaces, target module, required access, defense type, and validation environment.

Attack surfaces may be multi-label, but we require a distinction between the \textbf{primary entry surface} and \textbf{propagation/target surfaces}. If a malicious user prompt changes task planning and eventually produces a dangerous API call, AS02 is the primary entry, while AS08 and AS09 are downstream targets. If the attacker directly modifies a robot skill or action token, AS09 becomes the primary surface. For backdoors, malicious capability implanted during training is coded as AS01 even if deployment-time triggers are linguistic or visual. These rules prevent the final physical error from being indiscriminately relabeled as AS12.

\subsection{Six coding dimensions}
We use six complementary dimensions when deciding the primary surface:
\begin{enumerate}
  \item \textbf{Causal location:} locate the first polluted node along input/perception $\rightarrow$ world state $\rightarrow$ reasoning $\rightarrow$ planning $\rightarrow$ action interface $\rightarrow$ middleware $\rightarrow$ execution $\rightarrow$ multi-agent collaboration;
  \item \textbf{Representation type:} distinguish natural-language tasks, raw image/audio/LiDAR, structured state, CoT/hidden state, task plans, code/JSON/action tokens, and ROS messages;
  \item \textbf{Lifecycle stage:} distinguish development-time threats involving training, fine-tuning, and model distribution from deployment-time prompt, sensor, state, and network attacks;
  \item \textbf{Source and authority:} determine whether information originates from an authenticated user, the environment, RAG/memory, a tool, a cloud service, or another robot;
  \item \textbf{Propagation scope:} distinguish intra-robot compromise from robot-to-robot, shared-memory, and fleet-level propagation;
  \item \textbf{Security property:} treat integrity, availability, confidentiality/privacy, and physical safety as attack objectives rather than attack surfaces.
\end{enumerate}

\subsection{Methodological limitations}
Three limitations are unavoidable. First, some recent papers remain preprints and their venue status or experimental details may change. Second, multi-label coding necessarily involves judgment, especially when end-to-end VLAs tightly couple perception, reasoning, and action generation. Third, paper counts measure research attention rather than real-world attack probability; a heavily studied surface is not automatically more dangerous in deployment. The quantitative analysis should therefore be interpreted as a map of research density and attack--defense imbalance, not as an empirical incident-frequency estimate.

\section{Closed-Loop Security Model for Foundation-Model-Powered Embodied Agents}

\subsection{Unified information-flow abstraction}
Let $u$ denote the user task, $m_t$ contextual history, long-term memory, and retrieved information, $o_t$ raw multimodal observation, $s_t$ structured world state, $h_t$ an internal reasoning representation, $p_t$ a high-level task plan, $a_t$ an executable action representation, $r_t$ the robot execution state, and $x_t$ the physical environment state. A generic embodied loop can be written as
\begin{equation}
(u,m_t,o_t)\rightarrow s_t\rightarrow h_t\rightarrow p_t\rightarrow a_t\rightarrow r_{t+1}\rightarrow x_{t+1},
\end{equation}
with the next observation $o_{t+1}$ and feedback returning to the loop. In an end-to-end VLA, $s_t$, $h_t$, and $p_t$ may not be explicitly exposed. A world-model-based agent may additionally predict future trajectories
\begin{equation}
\hat{x}_{t+1:t+H}=W(s_t,a_{t:t+H-1}),
\end{equation}
and choose actions using predicted reward, safety cost, or affordance. Security analysis does not require all systems to expose identical modules; instead, it identifies the trust boundary created whenever one representation crosses into a more privileged decision or execution domain.

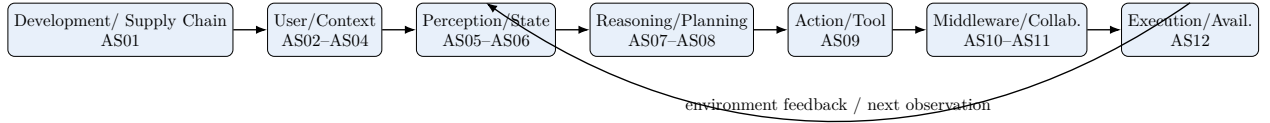
\begin{figure}[H]
\centering
\resizebox{0.98\textwidth}{!}{%
\begin{tikzpicture}[node distance=0.7cm, >=Latex, every node/.style={font=\small}]
\tikzstyle{box}=[draw,rounded corners,minimum height=1.1cm,minimum width=2.15cm,align=center,fill=softblue]
\node[box] (dev) {Development/\ Supply Chain\\AS01};
\node[box,right=of dev] (input) {User/Context\\AS02--AS04};
\node[box,right=of input] (sense) {Perception/State\\AS05--AS06};
\node[box,right=of sense] (cog) {Reasoning/Planning\\AS07--AS08};
\node[box,right=of cog] (act) {Action/Tool\\AS09};
\node[box,right=of act] (infra) {Middleware/Collab.\\AS10--AS11};
\node[box,right=of infra] (exec) {Execution/Avail.\\AS12};
\draw[->,thick] (dev)--(input); \draw[->,thick] (input)--(sense); \draw[->,thick] (sense)--(cog);
\draw[->,thick] (cog)--(act); \draw[->,thick] (act)--(infra); \draw[->,thick] (infra)--(exec);
\draw[->,thick,bend left=35] (exec.north) to node[above]{environment feedback / next observation} (sense.north);
\end{tikzpicture}}
\caption{Closed-loop trust boundaries in foundation-model-powered embodied agents. The primary attack surface is coded by the first directly compromised boundary; downstream modules are recorded as propagation or effect surfaces.}
\label{fig:loop}
\end{figure}

\subsection{Entry, propagation, target, and consequence}
For an attack $\mathcal{A}$, we use the tuple
\begin{equation}
\mathcal{A}=\langle B_0,M,P,E\rangle,
\end{equation}
where $B_0$ is the first compromised trust boundary, $M$ the attack mechanism, $P=(B_1,\ldots,B_k)$ the propagation path, and $E$ the final security effect. Consider physical prompt injection: if an adversary places malicious text in the environment and the vision/OCR stack correctly reads it, but the model incorrectly grants it command authority, then $B_0=$AS04. The effect may subsequently pass through visual-semantic encoding, AS08 planning, and AS09 action generation before causing a physical outcome. By contrast, if the adversary directly perturbs pixels, light, audio, or another raw signal, the first untrusted quantity is the sensor input and the primary surface is AS05.

This distinction prevents classification by final outcome. A robot executing a dangerous action because of prompt injection does not make AS12 the primary surface. AS12 is primary only when the attacker directly manipulates trajectories, actuator commands, halt/freeze logic, or an availability boundary. Likewise, a malicious prompt that causes an incorrect CoT does not automatically become an AS07 attack; AS07 is primary when reasoning or an internal representation itself is directly targeted.

\subsection{Adversary capabilities and access models}
The literature spans black-box to white-box adversaries. A \textbf{black-box adversary} can provide language, manipulate environmental objects, or query outputs. A \textbf{gray-box adversary} may know architecture, skill interfaces, or partial state without changing all parameters. A \textbf{white-box adversary} can access model weights, gradients, training data, or internal representations. A \textbf{system-level adversary} may additionally reach ROS topics, network links, credentials, cloud APIs, or inter-robot messages. Physical proximity is another important dimension because patches, printed prompts, 3D textures, optical signals, and acoustic injection depend on deployment conditions.

\subsection{Security, safety, and privacy}
We use \emph{security} for protection against intentional adversaries targeting integrity, availability, confidentiality/privacy, or physical safety, while \emph{safety} also includes failures caused by benign perception error, planning mistakes, dynamics uncertainty, or control instability. The final physical consequence can be identical even when causality differs. A naturally misperceived obstacle is a safety failure; an adversarial patch deliberately causing the same miss is a security-induced safety failure. VLALeaks further demonstrates that embodied models may expose membership information and other privacy properties\cite{A042}. A mature framework should therefore record adversary, first boundary, security objective, and physical consequence separately.

\section{A Five-Layer, Twelve-Surface Taxonomy}

\subsection{Five-layer organization}
For readability, we aggregate the twelve fine-grained surfaces into five layers. The \textbf{development and supply-chain layer} contains AS01 and asks whether malicious capability is encoded before deployment. The \textbf{external information-input layer} contains AS02--AS06 and covers authenticated user tasks, auxiliary context, the physical semantic environment, raw multimodal observations, and structured world state. The \textbf{cognition and decision layer} contains AS07--AS09 and covers internal reasoning, high-level planning, and executable action/tool representations. The \textbf{infrastructure and collaboration layer} contains AS10--AS11 and covers ROS/network/external services and cross-robot trust. Finally, the \textbf{physical execution and availability layer} contains AS12 and covers direct trajectory, control, freezing, and denial-of-service manipulation.

\begin{table}[H]
\centering\small
\caption{Definitions, representative objects, and boundary criteria for the twelve attack surfaces.}
\label{tab:surfaces}
\begin{tabularx}{\textwidth}{p{0.8cm}p{3.0cm}X p{3.6cm}}
\toprule
Code & Attack surface & Core definition & Representative carriers/objects\\
\midrule
AS01 & Model, data, and supply chain & The attack first occurs before deployment in a learning asset or model/component boundary. Malicious influence is encoded through training, fine-tuning, adapters, checkpoints, or third-party components and is later activated during deployment. & Training data, fine-tuning examples, LoRA/Adapter, checkpoints, third-party FM components \\
AS02 & User instruction and natural-language control & Malicious input enters through the channel explicitly treated as the current user's task or control intent. The first corrupted representation is task intent rather than environmental observation. & Text tasks, speech commands, adversarial suffixes, direct prompts \\
AS03 & Context, memory, and ICL/RAG & The attack targets runtime language context other than the current user instruction, including demonstrations, history, persistent memory, retrieved documents, tool returns, or cached plans. & ICL examples, memory, RAG, history, tool results \\
AS04 & Physical semantic environment & The adversary changes real semantic content that is correctly perceived by the robot, such as signs, displays, labels, QR codes, or instruction-like objects. The sensor may function correctly; the failure is semantic authority. & Printed text, signs, screens, labels, QR codes, scene text \\
AS05 & Multimodal perception and sensors & The first corruption occurs in raw observations or sensor signals, including pixels, physical/digital patches, 3D textures, optical signals, audio, or LiDAR. & Pixels, patches, 3D textures, cameras, audio, LiDAR \\
AS06 & World state, localization, and proprioception & The attack targets structured state consumed by planning or policy modules. Raw sensing can be intact while object relations, occupancy, localization, joint state, affordances, feedback, or summaries are forged. & Object relations, occupancy, pose, joint state, affordance, feedback \\
AS07 & Reasoning, CoT, and internal representations & The attack directly targets an explicit or latent intermediate reasoning process, such as CoT, hidden representations, attention, or cross-modal alignment, after input/state but before explicit task plans. & CoT, hidden state, attention, cross-modal alignment \\
AS08 & Task planning, goals, and policy intent & The adversary directly manipulates goals, task decomposition, action order, paths, recovery strategies, or high-level policy intent before concrete API/code/action encoding. & Goals, task decomposition, plans, paths, high-level policy intent \\
AS09 & Action, code, tool, and skill interfaces & The attack targets the structured interface from cognition to execution, where content becomes executable control representation capable of invoking physical privileges. & Code, JSON, API/tool calls, skills, action tokens/chunks \\
AS10 & Middleware, network, and external services & The attack occurs in robot infrastructure, including ROS/ROS2, DDS, topics/services, drivers, networks, cloud models, or remote APIs. The core issue is message, identity, credential, dependency, or transport security. & ROS/ROS2, DDS, topics/services, network, cloud model/API \\
AS11 & Multi-agent and multi-robot communication & Malicious or false information is supplied directly by another embodied agent/robot or shared collaboration state and propagates across subject boundaries. & Robot-to-robot messages, shared prompts, shared memory, claims \\
AS12 & Execution control and availability & The adversary directly manipulates physical execution, trajectories, stopping/freezing, safety shutdown, or resource availability. Upstream attacks that merely cause wrong actions are not automatically AS12. & Trajectories, actuator commands, halt/freeze, DoS, safety stop \\
\bottomrule
\end{tabularx}
\end{table}

\subsection{Why attack surfaces and attack mechanisms must be orthogonal}
Attack mechanisms describe \emph{how} a compromise is achieved; attack surfaces describe \emph{where} it first enters. A backdoor can be implanted in training data, a perception model, a state representation, or an action generator. AS05 can host adversarial patches, sensor spoofing, visual triggers, or deployment-time backdoor triggers. We therefore avoid labels such as ``jailbreak surface'' or ``backdoor surface'' and instead use mechanism as a second coding dimension.

The separation also handles composite attacks. TRAP uses a physical adversarial patch but targets VLA CoT reasoning; AS05 can be recorded as the input surface and AS07 as a critical internal target\cite{A031}. RIPA spans OCR, audio/STT, and LiDAR/state-vector channels, with the primary surface determined by whether the earliest corruption is a raw sensor or a structured state\cite{A038}. \emph{When Prompts Control Robots} shows that AS04 environmental/indirect injection and AS11 cross-robot propagation may occur in the same attack chain\cite{A043}.

\section{Attack Research: From the First Compromised Boundary to Physical Consequences}
This section reviews the attack literature using the twelve surfaces. For each surface we ask: what representation is first controlled by the adversary; how representative mechanisms operate; how effects propagate into plans and actions; and what major limitations remain.

\subsection{AS01: Model, data, and supply-chain attacks}
AS01 covers training data, fine-tuning examples, LoRA/Adapters, checkpoints, third-party perception/policy models, and dependent components. When malicious capability is encoded before deployment, the primary surface remains AS01 even if the runtime trigger arrives through language, vision, or state. This separation between \emph{implantation surface} and \emph{trigger carrier} prevents every visual-trigger backdoor from being misclassified as a perception attack.

BALD is a representative study of backdoors in embodied LLM decision making. It designs word, scenario, and knowledge triggers that preserve clean behavior but cause attacker-desired plans when selected textual or environmental conditions occur\cite{A003}. Later work extends backdoors to VLMs and VLAs. TrojanRobot/Robot Collapse studies perception-model and supply-chain compromise\cite{A008}; BEAT learns visual triggers for multimodal embodied decisions\cite{A010}; and BadVLA, DropVLA, Clean-Action Backdoor, and SilentDrift move the target from high-level decisions into action-level backdoors, sequential error accumulation, and action-chunk generation\cite{A013,A016,A017,A028}. INFUSE further emphasizes persistence after downstream fine-tuning\cite{A057}.

Two trends are visible. First, embodied backdoors are moving from conspicuous linguistic triggers and obviously unsafe outputs toward environmental conditions, physical-object triggers, and subtle action drift. Such attacks are substantially harder to catch with harmful-text filters. Second, end-to-end VLAs couple perception and action generation within one model, allowing training-time objectives to embed malicious behavior directly in continuous-control distributions without ever producing a human-auditable language plan.

The central defensive challenge is that clean-task performance does not establish trustworthiness. High-quality backdoors are explicitly optimized to preserve benign success. Standard benchmarks and a small number of safety prompts are therefore insufficient. Stronger guarantees require data provenance, weight/adapter integrity, model-differential analysis, trigger search, and persistence testing under downstream fine-tuning.

\subsection{AS02: User instruction and natural-language control}
AS02 is the authorized user-task channel. An adversary changes the system's intended task through malicious natural language, adversarial suffixes, jailbreaks, direct prompt injection, or speech commands. Its defining distinction from AS03 and AS04 is that the malicious text arrives through the interface explicitly granted authority to specify the current task.

\emph{On the Vulnerability of LLM/VLM-Controlled Robotics} and EIRAD showed early that linguistic perturbations and decision-level adversarial inputs can alter high-level embodied behavior\cite{A001,A002}. BadRobot extended jailbreak analysis from unsafe text to physical action and emphasized language--action safety misalignment\cite{A004}. RoboPAIR automated the search for prompts that induce simulated or real robots to execute unsafe behavior under multiple access assumptions\cite{A007}. POEX focused on \emph{policy-executable} jailbreaks, where the attack must not only obtain linguistic compliance but also produce a policy supported by robot skills\cite{A009}. BadNAVer and PINA extend related attacks to vision--language navigation\cite{A011,A037}.

These studies expose at least three success conditions: the model accepts the malicious intent, the planner produces a goal-consistent plan, and the executor can realize it. Conventional jailbreak success rate usually captures only the first. Many text-level attacks fail physically because required actions are unreachable, collision constraints intervene, skills are missing, or state preconditions are unsatisfied. Treating ``model did not refuse'' as physical ASR therefore systematically overestimates embodied risk.

At the same time, AS02 is easy to overemphasize. As robots increasingly depend on VLM perception, RAG, persistent memory, and structured world states, an attacker may not need to control the authenticated user at all. Future work should compare direct prompt attacks, indirect prompts, state corruption, and action manipulation under a common closed-loop protocol instead of repeatedly optimizing text-only jailbreak templates.

\subsection{AS03: Context, memory, and ICL/RAG}
AS03 includes in-context demonstrations, conversation history, long-term memory, retrieved documents, external tool returns, and cached plans. The core risk is not a malicious command from the current user but excessive trust in auxiliary evidence that may persist across rounds.

\emph{Compromising LLM Driven Embodied Agents with Contextual Backdoor Attacks} shows that poisoning a small number of in-context demonstrations can induce defective programs or plans under specific textual or visual conditions without changing base-model parameters\cite{A006}. CrossInject demonstrates that untrusted context can enter a multimodal agent through different modalities and create authority confusion during decision making\cite{A025}.

The importance of AS03 is likely to increase as embodied agents acquire persistent memory. A one-shot prompt injection often ends when the input disappears. Memory poisoning can be repeatedly retrieved as ``experience'', and automatic summarization, reflection, or skill caching may compress and entrench the malicious state. This resembles persistent compromise: brief access to one context channel may influence many future tasks.

Only 2 of 58 attack records in the current corpus directly cover AS03, making it one of the sparsest surfaces. This research density is misaligned with the trajectory of agent architectures. Important topics include memory provenance, RAG-document trust, tool-return integrity, experience-replay poisoning, secure memory write/read policies, and cross-task persistence metrics.

\subsection{AS04: Physical semantic environment}
AS04 is a distinctive embodied boundary. The adversary changes semantic content that physically exists in the scene and is correctly perceived by the robot---printed text, displays, signs, labels, QR codes, or instruction-like objects. The sensor can operate perfectly; the vulnerability is that environmental information is promoted from descriptive evidence to control authority.

\emph{The Shawshank Redemption of Embodied AI} systematizes indirect environmental jailbreaks\cite{A022}; CHAI frames the problem as command hijacking\cite{A024}; PI3D extends prompt injection into 3D environments\cite{A026}; and \emph{Hijacking Robots with a Piece of Paper} studies real-world physical carriers for VLM-controlled robots\cite{A046}. More recent work on trust-boundary confusion emphasizes that correct scene-text recognition can still produce unsafe behavior when source and authority are conflated\cite{A058}.

The distinction between AS04 and AS05 is critical. If an adversary places real text in the scene and the camera/OCR stack reads it correctly, the first compromised boundary is semantic authority and the attack is AS04. If the adversary perturbs pixels, light, or another signal such that the raw observation itself is corrupted, the primary surface is AS05. The defenses differ correspondingly: AS04 requires provenance and authority control, whereas AS05 requires perceptual robustness.

Simple malicious-text classification does not solve AS04. A sentence may be perfectly benign yet should not be allowed to override the authenticated user goal. The core problem is \emph{semantic authority}: a robot must know not only what environmental content means, but also what that content is permitted to modify, which skills it may invoke, and whether it can alter safety rules.

\subsection{AS05: Multimodal perception and sensors}
AS05 is one of the most heavily studied surfaces, appearing in 29 of 58 attack records. The first compromised object is a raw observation or sensor signal---image pixels, physical/digital patches, 3D textures, optical signals, audio, LiDAR, or related channels. Even if the observation is later converted into text or structured state, coding follows the earliest distorted sensor boundary.

The rise of VLA models has accelerated this area. Human/Model-Agnostic Attacks, AttackVLA, and VLA-Fool investigate model-agnostic, benchmark-oriented, and cross-modal adversarial robustness, respectively\cite{A014,A018,A019}. UPA-RFAS, Partially Observable Patch, Tex3D, and PhysPatch emphasize physical patches, partial observability, 3D textures, and physical transferability\cite{A020,A040,A032,A056}. Phantom Menace directly considers physical sensor attacks against VLAs\cite{A039}, while DRIFT and DURA move the objective from single-step output error toward trajectory derailment and unrestricted robotic attacks\cite{A048,A050}.

VLA perception attacks differ fundamentally from attacks on static image classifiers because errors can be temporally amplified. A visual perturbation changes the start of an action chunk; the action then changes viewpoint and state; the new observation feeds back into the next action. This creates a closed feedback loop of observation error, action error, and new observation error. Pixel perturbation magnitude alone is therefore a poor proxy for embodied risk. Evaluation should include trajectory deviation, task-stage misalignment, collision, safety-predicate violations, and final environment state.

Although AS05 is heavily studied, real-world validation remains weaker than digital evaluation. Physical attacks must survive changes in viewing angle, distance, illumination, occlusion, robot motion, and automatic camera exposure. Strong future studies should report physical realizability, cross-robot transfer, long-horizon persistence, and defense latency in addition to offline output changes.

\subsection{AS06: World state, localization, and proprioception}
AS06 targets structured state used by planners or policies: object relations, scene graphs, occupancy, localization, joint state, affordances, execution feedback, and state summaries. Raw sensing may be correct while the representation consumed by the planner is forged, corrupted, or used as a trigger.

BALD's scenario and knowledge triggers already show that environment state can serve as a backdoor condition\cite{A003}. State Backdoor explicitly studies stealthy poisoning in VLA state space\cite{A027}. RIPA brings OCR, audio/STT, and LiDAR/state vectors into the ROS2 prompt-injection pipeline\cite{A038}. \emph{When World Models Dream Wrong} extends this concern into physically conditioned state and prediction processes inside world models\cite{A041}.

The security importance of AS06 lies in the distinction between \emph{fact integrity} and \emph{instruction integrity}. An attacker does not need to tell the robot to ``go to the wrong location''. If object position, reachability, affordance, or execution feedback is forged to align with the attacker's goal, the planner may remain perfectly faithful to the original user instruction while autonomously deriving an incorrect plan. Such evidence can appear more benign than an explicit jailbreak and may evade content-focused defenses.

Only 4 of 58 attack records cover AS06 directly, even though world models, explicit scene representations, and proprioception are becoming increasingly important. Future work should examine state provenance, sensor-to-state consistency, temporal consistency, feedback spoofing, affordance corruption, and predictive safety illusions in world models.

\subsection{AS07: Reasoning, CoT, and internal representations}
AS07 lies between external input/world state and explicit task planning and covers chain-of-thought, latent representations, attention, cross-modal alignment, and intermediate causal judgments. If an incorrect CoT is merely the downstream result of a malicious prompt, AS07 is normally an effect surface. It becomes primary when reasoning or an internal representation is directly targeted.

\emph{Altered Thoughts, Altered Actions} probes CoT vulnerabilities in VLA manipulation\cite{A030}. TRAP uses adversarial patches to hijack CoT reasoning, illustrating AS05 as the input surface and AS07 as an internal target\cite{A031}. FlowHijack attacks the generation dynamics of flow-matching VLAs\cite{A034}; JailWAM and world-model attacks show that internal prediction and action reasoning can themselves become integrity targets\cite{A036,A041}. VLAGuard/VASA and AGSD-related attacks further expose weaknesses in attention and spatial alignment\cite{A047,A049}.

A fundamental challenge is observability. Unlike prompts or API calls, hidden states and attention are not stable external interfaces and may not have directly interpretable semantics. Both attack success and defense effectiveness are therefore harder to define. Yet end-to-end VLAs increasingly eliminate explicit plans, making internal representations one of the few places where the system can be inspected before an action is emitted.

Future work should distinguish carefully between an input attack that incidentally changes internal representations and an attack that explicitly optimizes those representations as the target. Causal attribution, representation anomaly detection, attention consistency, and mechanistic evaluation are needed to prevent AS07 from becoming an unfalsifiable catch-all category.

\subsection{AS08: Task planning, goals, and policy intent}
AS08 captures integrity of goals, task decomposition, plans, paths, recovery strategies, and high-level policy intent. If the adversary directly modifies these objects, AS08 is the entry surface; if planning changes only because of an upstream prompt, sensor, or state attack, AS08 is a propagation or target surface.

EIRAD focuses on decision-level embodied models\cite{A002}. BadRobot, RoboPAIR, and POEX all treat the generation of an executable malicious plan as a critical stage beyond language compliance\cite{A004,A007,A009}. BadNAVer and PINA extend plan/path hijacking to navigation\cite{A011,A037}. Environmental jailbreaks and CHAI show that physical semantic content can enter through perception and alter task intent\cite{A022,A024}. Blindfold demonstrates the tight coupling between action-level manipulation and high-level agent planning\cite{A029}.

AS08 is especially important for hierarchical robot architectures, where an LLM produces a symbolic plan or program that is later executed by a conventional controller. The architecture naturally provides a security checkpoint before the plan acquires physical authority. However, if planner output is trusted by default, prompt, state, and perception attacks all converge at this boundary.

Evaluation should not treat any textual plan difference as attack success. More meaningful criteria include whether goal predicates change, whether critical subtasks are substituted, whether action order satisfies the attacker's objective, whether the plan is executable, and whether execution produces the intended final-state deviation.

\subsection{AS09: Action, code, tool, and skill interfaces}
AS09 is tied with AS05 for the highest attack coverage, appearing in 29 of 58 attack records. It occupies the boundary between cognition and physical control and includes code, JSON, function/tool calls, robot skills, action tokens, continuous actions, and action chunks. It is effectively the \textbf{privilege boundary at which model semantics acquire physical execution authority}.

BadRobot and RoboPAIR both emphasize that malicious intent matters only if it is translated into executable behavior\cite{A004,A007}. POEX explicitly targets policy-executable jailbreaks\cite{A009}. \emph{Adversarial Attacks on Robotic VLA Models}, DropVLA, Clean-Action Backdoor, and SilentDrift push attacks directly into action spaces, action-level backdoors, and action chunks\cite{A012,A016,A017,A028}. Blindfold, FlowHijack, DRIFT, and DURA further show that high-level language can remain seemingly reasonable while action proxies, generation dynamics, or continuous trajectories are corrupted\cite{A029,A034,A048,A050}.

The density of AS09 research reflects a central lesson: filtering natural language is not sufficient. A model may provide a benign explanation while function parameters, coordinates, velocity, grasp targets, or action chunks have already changed. For end-to-end VLAs, there may be no natural-language intermediate plan at all, making the action space the only enforceable privilege boundary.

Defenses should therefore include least-privilege skill interfaces, action-schema validation, parameter bounds, reachability/CBF checks, and independent runtime monitors. It is also important to distinguish action representation from low-level execution: direct manipulation of API/skill/action tokens is AS09, whereas direct manipulation of actuator commands or trajectory executors is AS12.

\subsection{AS10: Middleware, network, and external services}
AS10 covers ROS/ROS2, DDS, topics/services, drivers, network links, cloud-hosted LLM/VLM services, credentials, and remote APIs. The core problem is not that the model ``understands incorrectly'', but that system software, message transport, identity, or service dependencies lose confidentiality, integrity, or availability.

\emph{From Prompts to Motors} considers man-in-the-middle risks in LLM-connected robotic systems\cite{A051}; Net-GPT is a representative example of LLM-assisted MITM for unmanned systems\cite{A052}; and work on supply-chain exploitation and the insecurity of Secure ROS2 demonstrates that keystores, configuration, and deployment can remain attack surfaces even when ROS2 security features are enabled\cite{A053,A054}.

Foundation models can amplify AS10. Historically, a corrupted ROS topic might affect one control node. In a foundation-model-powered system, the same message may be serialized into text or state and fed to an LLM/VLM as trusted evidence for complex reasoning. Cloud inference additionally introduces API credentials, endpoint availability, model substitution, and remote dependency risks.

Only 6 of 58 attack records and 2 of 61 defense records cover AS10, making it a pronounced weak spot. Future research should connect traditional robotics cybersecurity with agent security through message signing, device identity, cloud-model authentication, endpoint attestation, state provenance, and least-privilege middleware policies.

\subsection{AS11: Multi-agent and multi-robot communication}
The defining property of AS11 is trust across principals. Malicious data comes directly from another robot/agent, a shared prompt, shared memory, task-allocation message, state claim, or collaboration protocol. Even if the message eventually enters an LLM as plain text, the primary boundary is robot-to-robot trust.

\emph{When Prompts Control Robots} shows how prompt injection can propagate in multi-agent robotic systems\cite{A043}. InfectBot studies how compromise of one robot can propagate unsafe actions\cite{A044}. \emph{When Coordination Becomes a Threat} directly examines communication attacks in LLM-controlled multi-robot systems\cite{A045}, while \emph{Automated Discovery of Semantic Attacks in Multi-Robot Navigation Systems} studies automated discovery of such semantic attacks\cite{A055}.

Unlike single-agent memory poisoning, multi-robot attacks introduce explicit source identity and authorization. A shared claim must not only be plausible; the receiver must establish who issued it, whether that robot has authority to make the claim, and whether independent evidence supports it. If a fleet planner treats peers as trusted sensors, compromise of one robot creates a lateral-propagation channel.

The literature remains sparse, but AS11 is likely to become more important as fleet-level coordination and shared foundation-model services expand. Promising directions include Byzantine-resilient coordination, claim provenance, cross-agent isolation, trust scoring, quorum verification, and benchmarks for infection radius and propagation depth.

\subsection{AS12: Execution control and availability}
AS12 is primary only when an adversary directly acts on low-level execution, trajectories, actuator commands, halt/freeze behavior, safety shutdown, or runtime resources. An upstream attack that merely causes the robot to choose a wrong action does not automatically become AS12.

FreezeVLA shows that adversarial input can induce action freezing and thereby create an availability failure\cite{A021}. Semantic Denial of Service exploits semantic or safety triggers to make an LLM-controlled robot repeatedly refuse or stop\cite{A035}. SilentDrift illustrates how subtle continuous-action drift can accumulate over long horizons\cite{A028}. Collectively, these works show that an embodied attack need not produce an obviously hazardous action; rendering the robot unavailable at a critical time can be equally consequential.

AS12 has relatively few direct attacks but many defenses, reflecting an engineering preference to use execution as a final safety barrier. This is sensible but insufficient: downstream monitors may detect danger only after upstream compromise has already degraded availability, and single-step action checks may miss long-horizon risk. Execution defense should therefore be combined with provenance, state integrity, and plan verification.

\section{The Two-Dimensional Relationship Between Attack Mechanisms and Attack Surfaces}

\subsection{Mechanisms are not surfaces}
The major mechanisms in the corpus include jailbreak/direct prompt injection, indirect/physical prompt injection, backdoors, data/context poisoning, adversarial patches/perturbations, sensor spoofing, reasoning/attention hijacking, action manipulation, MITM/middleware compromise, multi-agent propagation, and semantic denial of service. Table~\ref{tab:mechanism-surface} summarizes their typical relationship to primary attack surfaces.

\begin{table}[H]
\centering\small
\caption{Attack mechanisms and their common primary attack surfaces.}
\label{tab:mechanism-surface}
\begin{tabularx}{\textwidth}{p{4.1cm}p{3.1cm}X}
\toprule
Attack mechanism & Common primary surfaces & Decision rule\\
\midrule
Jailbreak / Direct Prompt Injection & AS02; sometimes AS08/AS09 & If malicious content comes directly from the currently authorized user-task channel, AS02 is the primary entry; plan/action changes are normally downstream targets.\\
Indirect / Physical Prompt Injection & AS03 / AS04 / AS05 & Depends on whether untrusted language comes from auxiliary context, real environmental semantics, or a corrupted raw sensor.\\
Backdoor / Poisoning & AS01 & Malicious capability is implanted during training, fine-tuning, or model distribution; runtime triggers are recorded separately.\\
Adversarial Patch / Sensor Spoofing & AS05 & The first distorted quantity is a raw image, audio, LiDAR, optical, or related sensor signal.\\
State Manipulation & AS06 & The adversary directly changes object relations, occupancy, pose, joint state, affordances, feedback, or other structured state.\\
Reasoning / Attention Hijacking & AS07 & The adversary directly targets CoT, attention, hidden representations, or cross-modal alignment.\\
Action / Tool Manipulation & AS09 & The adversary directly changes code, API calls, skills, action tokens/chunks, or continuous action representations.\\
MITM / Middleware Compromise & AS10 & The attack targets ROS/DDS, networks, credentials, cloud services, or message identity.\\
Multi-Agent Propagation & AS11 & Malicious information directly originates from another robot/agent or a shared collaboration state.\\
DoS / Freezing & AS12 & AS12 is primary when the adversary directly controls a halt/freeze/trajectory/availability boundary.\\
\bottomrule
\end{tabularx}
\end{table}

\subsection{Composite attack chains}
Real attacks are often multi-stage rather than single-point events. We recommend representing them as entry--propagation--consequence chains. For example,
\begin{equation}
\text{AS04 environmental text}\rightarrow\text{AS05 perception}\rightarrow\text{AS08 plan}\rightarrow\text{AS09 action}\rightarrow\text{physical effect},
\end{equation}
or
\begin{equation}
\text{AS01 poisoned VLA}\xrightarrow{\text{visual trigger}}\text{AS05}\rightarrow\text{AS09 malicious action chunk}\rightarrow\text{AS12 effect}.
\end{equation}
The first attack has AS04 as its primary surface; the second has AS01 as its primary implantation surface, although AS05 may also be recorded as a critical deployment-time trigger surface. Multi-label coding preserves the causal chain without giving up the requirement to identify a primary entry boundary.

\section{Defense Research: From Point Filters to Cross-Layer Security Architectures}
We do not organize defenses by mechanically repeating the twelve attack surfaces. Instead, defenses are grouped by where they intervene, what trusted information they depend on, and whether they can provide runtime guarantees. A single defense may cover several surfaces.

\subsection{Training-time and post-training alignment and model hardening}
The first family strengthens the model before deployment through safety alignment, robust fine-tuning, reinforcement post-training, parameter merging, unlearning, inductive bias, and backdoor erasure. SafeVLA incorporates safety objectives through constrained learning\cite{D031}; RobustVLA uses robustness-aware reinforcement post-training\cite{D040}; RETAIN and MergeVLA use parameter/model merging to improve robustness\cite{D041,D042}; VLA-Forget studies unlearning for embodied foundation models\cite{D039}; and \emph{Towards Safe Robot Foundation Models Using Inductive Biases} explores architectural and inductive-bias mechanisms\cite{D061}.

Backdoor-specific defenses increasingly operate at a mechanistic level. TrustVLA uses causal footprints, trigger localization, and inpainting at inference time\cite{D045}; Bera detects abnormal attention and reconstructs suspicious visual tokens\cite{D058}; and VLAGuard/APFT together with SARF strengthen models through attention-protective and structure-aware robust fine-tuning\cite{D046,D047}.

The advantage of model hardening is low runtime overhead and the possibility of improving average robustness across many tasks. The main limitation is that deployment contexts and attack distributions remain open-ended. Closed models and API-only systems may not expose weights or training access at all. Consequently, model hardening is best treated as a foundational layer rather than a complete security solution.

\subsection{Input filtering, provenance validation, and privilege separation}
The second family acts before untrusted language, environmental text, or multimodal content reaches the central planner. SafeEmbodAI combines secure prompting, state management, and safety validation against prompt injection in mobile robots\cite{D003}; J-DAPT uses a multimodal jailbreak detector with domain adaptation\cite{D010}; physical-prompt-injection mitigations combine prompt hardening, two-stage verification, and text masking\cite{D051}; and the RIPA Hybrid Semantic Firewall combines rules and semantic filtering across multimodal sensor-injection channels\cite{D050}.

The deeper challenge is not simply to decide whether a sentence is malicious. Embodied systems must decide whether the information is \emph{authorized to alter control flow}. Scene text saying ``turn left'', a RAG document saying ``ignore previous instructions'', or another robot claiming ``the target is safe'' may all be syntactically plausible. They should nevertheless have different privileges. Input security should therefore evolve from content filtering toward provenance-aware privilege control: source identity, authority, integrity, freshness, and the set of actions that each source may influence should accompany information into the planner.

Multi-agent defenses make this need more explicit. Per-Agent LLM Isolation separates contexts to prevent one agent's malicious prompt from directly contaminating another\cite{D052}. Trust-Boundary-Aware Multi-Agent Defense explicitly models trust between visual injection and multi-agent decision making\cite{D059}.

\subsection{World-state integrity, cross-modal consistency, and verifiable facts}
Defenses for AS06 are concerned less with filtering user language than with ensuring that the state consumed by the planner matches the physical world. RoboGuard contextualizes safety rules into temporal logic and combines them with system state\cite{D008}. RoboSafe uses executable safety logic, memory, and predictive/reflective reasoning to constrain planning and execution\cite{D012}. SafeGate combines task safety contracts with SMT/Z3-like verification before execution\cite{D014}.

For navigation and continuous environments, Safe-VLN supplements learned reasoning with occupancy-aware collision avoidance\cite{D025}. Affordance Field Intervention uses affordance-field rollback and recovery when manipulation falls into memory traps\cite{D026}. Causal Scene Narration combines structured scene descriptions with runtime intent and constraint monitoring\cite{D029}.

The long-term goal is a notion of \emph{fact-layer security}: object relations, pose, occupancy, affordances, joint states, and feedback should carry provenance and support cross-modal and temporal consistency checks. Otherwise, a plan verifier can be perfectly implemented yet still produce a predictive safety illusion if the state it verifies is itself forged.

\subsection{Internal representations, attention, and uncertainty intervention}
When end-to-end VLAs do not expose explicit plans, internal representations may become important defense objects. CEE performs hidden-state safety steering for embodied LLMs\cite{D009}; SAFE-Dict uses concept-based dictionary learning for inference-time intervention\cite{D044}; \emph{Ask Before You Act} uses token-level uncertainty to decide when human intervention is needed\cite{D022}; and SAFE uses latent features for multitask failure detection\cite{D016}.

These approaches can detect risk before unsafe actions are emitted and may cover failures that language filters miss. Their limitations include model-access requirements, uncertain semantic stability of latent features, limited cross-model transfer, and questions of causal validity. Such defenses are especially difficult for closed-source VLAs.

A central open issue is linking representation anomalies to actual physical risk. Does a hidden-state shift reliably imply unsafe behavior? Is attention concentration on a trigger causally responsible for action failure? Should high uncertainty always stop execution? Without closed-loop validation, internal metrics risk remaining offline proxies rather than operational guarantees.

\subsection{Plan verification, formal logic, and independent safety judgment}
Planning-layer defense is one of the most mature embodied-safety paradigms. Safety Chip translates natural-language safety requirements into Linear Temporal Logic and detects or prunes violating actions in candidate LLM plans\cite{D001}. SafePlan combines prompt sanity checks, invariants, and pre/postconditions with reasoning\cite{D006}. RoboGuard, SafeGate, and LogicGuard use temporal logic, contracts/SMT, and logic-based critics to strengthen plan verification\cite{D008,D014,D015}. Reachability-based work additionally incorporates system dynamics and reachable sets into formal guarantees\cite{D005}.

PROTEA uses an independent LLM-as-a-Judge or plan verifier between planner and executor\cite{D013}. Cross-Layer Sequence Supervision argues that safety supervision should follow the language--plan--action chain rather than occur once at the language boundary\cite{D002}. These approaches implement a form of privilege separation: a generative model may propose a behavior, but a distinct component grants execution authority.

The main bottleneck is specification coverage. Real-world safety rules are difficult to enumerate completely, and the verifier may depend on untrusted state. An LLM-based judge may also share vulnerabilities with the planner if both use the same model, prompt patterns, or state source. A promising direction is therefore to combine hard formal constraints, learned semantic judgment, and explicit state provenance rather than rely on any single verifier.

\subsection{Action-level runtime monitoring, correction, and recovery}
The most densely populated defense area is AS09/AS12. RoVer applies runtime verification to VLA policies\cite{D021}; SafeDec constrains decoding in autoregressive embodied policies\cite{D023}; VLSA adds a plug-and-play safety constraint layer\cite{D030}; and BYOVLA uses runtime observation intervention to improve visual robustness\cite{D024}. These systems place the final safety gate immediately before physical authority is exercised.

A complementary family monitors whether the robot has already entered or is about to enter a failure state. AHA detects and reasons about manipulation failures with a VLM\cite{D017}; Guardian uses an external VLM to monitor planning and execution errors\cite{D018}; Robot Success Detection, FPC-VLA, REFLECT, and FailSafe cover success/failure detection, prediction/correction, experience summarization, and recovery\cite{D019,D020,D027,D028}.

Runtime defense has an important model-agnostic property: even if prompt, perception, or planning defenses fail, the system can still stop an action that violates a safety predicate. The tradeoff is latency and conservatism. High-capacity LLM/VLM supervisors may be too slow for high-frequency control, while overly conservative monitors can create frequent false stops. Evaluation must therefore jointly report safety gain, clean task success, latency, intervention frequency, and recovery success.

\subsection{Middleware security, multi-robot isolation, and Byzantine resilience}
AS10 and AS11 remain the least developed defense areas. Structured output verification and schema defenses check the structure of commands that cross system interfaces\cite{D054}, while traditional MITM detectors monitor network integrity\cite{D057}. In multi-robot settings, CPV Gate verifies provenance and independently checks claims\cite{D053}; RoboRebound studies bounded-time resilience\cite{D055}; and decentralized blocklist protocols draw on Byzantine-fault-tolerance ideas for swarm-scale malicious nodes\cite{D056}.

This convergence shows that foundation-model-powered embodied security is increasingly overlapping with distributed-systems security. Future fleet-level systems require not only safe models but also device identity, message authentication, least privilege, shared-memory isolation, claim provenance, and Byzantine-resilient coordination. Otherwise, strong single-robot safeguards may still be bypassed through a compromised collaborator or middleware path.

\section{Quantitative Research Landscape and Attack--Defense Asymmetry}

\subsection{Time, model type, and inclusion tier}
Table~\ref{tab:year} shows the temporal distribution of the corpus. Attack research grows substantially after 2024; by August 15, 2026, we had collected 30 attack records from 2026 alone. Defense research also grows quickly in 2025--2026. Because 2026 is an incomplete year, these counts reflect research activity rather than a strict year-over-year growth rate.

\begin{table}[H]
\centering
\caption{Attack and defense records by year.}
\label{tab:year}
\begin{tabular}{lrrrrr}
\toprule
Category & 2022 & 2023 & 2024 & 2025 & 2026 (through Aug. 15)\\
\midrule
Attacks & 1 & 1 & 5 & 21 & 30\\
Defenses & 0 & 3 & 8 & 27 & 23\\
\bottomrule
\end{tabular}
\end{table}

By model type, VLA studies account for 25/58 attack records and LLM studies for 14/58; the remaining records span LLM/VLM, VLM/MLLM, world models/WAMs, and system-level work. In the defense corpus, VLA accounts for 24/61 and LLM for 16/61. The trend indicates a shift from the early question ``can an LLM planner be jailbroken?'' toward the robustness, action safety, and backdoor behavior of end-to-end VLAs. World-model/WAM security remains comparatively nascent.

\subsection{Coverage of the twelve attack surfaces}
Because one paper may cover several surfaces, the percentages in Table~\ref{tab:coverage} indicate the fraction of records that include each surface; they do not sum to 100\%.

\begin{table}[H]
\centering\small
\caption{Coverage of the twelve attack surfaces by attack and defense records.}
\label{tab:coverage}
\begin{tabular}{p{0.75cm}p{3.1cm}rrrrr}
\toprule
Code & Attack surface & Attack count & Attack coverage & Defense count & Defense coverage & D/A ratio\\
\midrule
AS01 & Model, data, and supply chain & 16 & 27.6\% & 12 & 19.7\% & 0.75 \\
AS02 & User instruction and language control & 15 & 25.9\% & 8 & 13.1\% & 0.53 \\
AS03 & Context, memory, and ICL/RAG & 2 & 3.4\% & 2 & 3.3\% & 1.00 \\
AS04 & Physical semantic environment & 10 & 17.2\% & 5 & 8.2\% & 0.50 \\
AS05 & Multimodal perception and sensors & 29 & 50.0\% & 20 & 32.8\% & 0.69 \\
AS06 & World state, localization, proprioception & 4 & 6.9\% & 8 & 13.1\% & 2.00 \\
AS07 & Reasoning, CoT, internal representation & 12 & 20.7\% & 10 & 16.4\% & 0.83 \\
AS08 & Task planning, goals, policy intent & 17 & 29.3\% & 22 & 36.1\% & 1.29 \\
AS09 & Action, code, tool, and skill interfaces & 29 & 50.0\% & 46 & 75.4\% & 1.59 \\
AS10 & Middleware, network, external services & 6 & 10.3\% & 2 & 3.3\% & 0.33 \\
AS11 & Multi-agent/multi-robot communication & 4 & 6.9\% & 4 & 6.6\% & 1.00 \\
AS12 & Execution control and availability & 5 & 8.6\% & 18 & 29.5\% & 3.60 \\
\bottomrule
\end{tabular}
\end{table}

\subsection{Observation 1: attack research forms a perception--action double peak}
AS05 and AS09 each appear in 29/58 attack records, or 50.0\%. This double peak mirrors the structure of VLA systems: visual input is the primary interface through which the model observes the world, while action output is the primary interface through which the model obtains physical authority. An adversary can therefore either corrupt perception at the input or bypass high-level semantics and manipulate action tokens, chunks, or trajectories directly. AS08 task planning (17/58), AS01 supply chain (16/58), and AS02 user instruction (15/58) form a second tier, reflecting the continuing importance of LLM-planner security and VLA backdoors.

\subsection{Observation 2: defenses shift downstream toward action and execution}
AS09 appears in 46/61 defense records (75.4\%), AS12 in 18/61 (29.5\%), and AS08 in 22/61 (36.1\%). The defense-to-attack coverage ratio is 1.59 for AS09 and 3.60 for AS12. This distribution reflects a clear engineering preference: when a foundation model cannot be assumed fully trustworthy, researchers place a final safety barrier at the plan-to-action and execution boundaries using action validators, runtime monitors, reachability checks, failure detectors, and safety shields.

This approach is important but cannot substitute for upstream integrity. A downstream monitor may decide whether ``this action is dangerous now'', yet fail to recognize persistent memory poisoning, forged world state, or a false claim from another robot. Exclusive reliance on AS09/AS12 defense can also raise false-stop rates and reduce task availability.

\subsection{Observation 3: AS03, AS10, and AS11 are pronounced sparse regions}
AS03 has only 2 attack and 2 defense records; AS10 has 6 attacks and 2 defenses; AS11 has 4 attacks and 4 defenses. AS10 has the lowest defense-to-attack ratio, 0.33. As embodied systems become more dependent on RAG, long-term memory, cloud inference, ROS2, and fleet coordination, these sparse regions may represent cases where the research structure is lagging behind the system architecture.

\subsection{Observation 4: AS06 has more defenses than direct attacks but lacks a unified state threat model}
AS06 appears in only 4 direct attack records but 8 defense records. One explanation is that plan verification, collision avoidance, and runtime monitoring naturally depend on world state and therefore indirectly protect AS06. Yet work explicitly targeting state poisoning, feedback spoofing, affordance corruption, and world-model state attacks remains limited. In other words, the literature more often \emph{uses state to make safety decisions} than studies \emph{how that state can itself be compromised}.

\section{Evaluation: From Model Responses to Closed-Loop Physical Consequences}

\subsection{Why text-level ASR is insufficient}
Traditional LLM jailbreak evaluation often treats a harmful response or refusal bypass as success. An embodied system contains at least four additional stages: malicious intent acceptance, planning deviation, executable action generation, and physical consequence. Let $N$ be the number of test tasks. A planning-layer attack success rate may be written as
\begin{equation}
\mathrm{P\text{-}ASR}=\frac{1}{N}\sum_{i=1}^{N}\mathbb{I}[p_i\models G_a],
\end{equation}
and an execution-layer attack success rate as
\begin{equation}
\mathrm{E\text{-}ASR}=\frac{1}{N}\sum_{i=1}^{N}\mathbb{I}[x_i^{\mathrm{final}}\models G_a],
\end{equation}
where $G_a$ is an adversarial goal predicate. Their difference captures a planning-to-execution transfer gap, while $\mathrm{E\text{-}ASR}/\mathrm{P\text{-}ASR}$ is a coarse transfer ratio. Even if a study uses different terminology, planner compromise and physical consequence should be reported separately.

\subsection{Recommended attack-evaluation dimensions}
A strong embodied-attack benchmark should report at least the following:
\begin{itemize}
  \item \textbf{Clean utility:} clean success or benign task completion, ensuring that an attack does not achieve apparent success merely by destroying all task performance;
  \item \textbf{Stage-wise ASR:} success at intent acceptance, planning deviation, action generation, and final-state consequence;
  \item \textbf{Stealthiness:} textual/visual/state perturbation magnitude, observability of action drift, trigger naturalness, and detector hit rate;
  \item \textbf{Transferability:} transfer across models, scenes, robots, and simulation-to-reality settings;
  \item \textbf{Persistence:} whether compromise survives long action chunks, memory updates, fine-tuning, or multi-agent propagation;
  \item \textbf{Physical realizability:} robustness to viewpoint, distance, lighting, occlusion, dynamic backgrounds, sensor noise, and real dynamics;
  \item \textbf{Cost and access:} query count, attacker privilege, physical deployment effort, and whether white-box gradients or supply-chain control are required.
\end{itemize}

\subsection{Defense evaluation cannot stop at detection rate}
Defense evaluation must jointly measure attack reduction and benign-task degradation. If attack success before and after defense is $ASR_0$ and $ASR_d$, respectively,
\begin{equation}
\Delta ASR=ASR_0-ASR_d,
\end{equation}
but clean-success degradation, false positive/false stop rate, runtime latency, intervention frequency, recovery success, and resource overhead are equally important. A defense that achieves near-zero ASR by always refusing to execute has little operational value; safety and utility must be evaluated together.

For LLM/VLM runtime supervisors, studies should also state whether the monitor can be attacked by the same source. If planner and judge use the same model family, prompt pattern, or corrupted world state, the ``independent'' defense may simply replicate the same vulnerability. Benchmarks should therefore record defense independence and the size of the trusted computing base.

\subsection{Simulation, real robots, and reproducibility}
A large fraction of the corpus is still evaluated primarily in simulation, although a growing subset uses mobile robots, quadrupeds, manipulation platforms, ROS2 systems, and real sensor conditions. Simulation offers controllable variables, large-scale repetition, and precise final-state predicates, but it often idealizes vision, dynamics, communication latency, and safety-stop behavior.

We recommend a three-level evaluation protocol. \textbf{Level 1} uses standard simulators for large, reproducible benchmarks. \textbf{Level 2} bridges digital perturbations to physical sensing and tests whether robustness survives realistic sensor transformations. \textbf{Level 3} evaluates attacks and defenses in a real closed-loop robot. High-risk physical tests should use safety cages, low-speed/low-force execution, soft targets, and independent emergency stop mechanisms so that security research does not itself create unnecessary hazards.

\subsection{What a unified benchmark should cover}
A unified benchmark should not consist only of prompt jailbreaks. At a minimum, the same task suite should cover benign tasks, direct malicious instructions, indirect/environmental injection, sensor perturbation, state corruption, memory/context poisoning, action/interface manipulation, and multi-agent propagation. Every attack family should be evaluated on the same clean-success subset, and logs should be preserved at model, planning, action, and physical layers. Only then can the relative risks of different attack surfaces be meaningfully compared.

\section{Open Problems and Future Research Directions}

\subsection{From point guardrails to compositional cross-layer defense}
Current systems often place a guardrail at one point---prompt, vision, plan, or action. Real attacks may cross several representations: environmental text becomes visual features, then language tokens, then a changed plan, and finally an action. If each defense lacks provenance, authority, and decision history, every local step may appear individually reasonable while the composition is unsafe. Future work should pursue compositional security guarantees that characterize which safety properties survive as untrusted data pass through sanitizers, planners, verifiers, and executors.

\subsection{Provenance and delegated authority should become first-class objects}
Many attacks are dangerous not because the content is intrinsically malicious, but because the content receives authority it should not have. Future robot agents should attach source, identity, authorization level, timestamp, integrity evidence, and permitted influence scope to each information item. Environmental text may inform scene understanding but should not override the authenticated user goal; a RAG document may supply knowledge but should not directly invoke a motor skill; another robot may submit a claim but should require independent verification before modifying shared world state. This direction can draw from information-flow control, capability systems, and zero-trust architectures.

\subsection{World-state integrity is likely to become a core next-generation boundary}
As scene graphs, world models, affordance maps, persistent state, and predictive simulators become standard, the security question shifts from ``was the model prompted maliciously?'' toward ``are the facts on which the model reasons trustworthy?'' Future work should systematically study state poisoning, feedback spoofing, dynamics corruption, affordance manipulation, and predictive safety illusions, while building sensor-to-state provenance and cross-modal consistency checks. A robot that rejects malicious prompts but fully trusts forged state remains insecure.

\subsection{Long-term memory turns transient attacks into persistent compromise}
Persistent memory, automatic reflection, experience summarization, and skill learning rewrite previous interactions into future context. If malicious content is stored, a short-lived injection can recur across many later tasks. Needed mechanisms include memory quarantine, write-time trust checks, read-time provenance, decay/expiry, secure summarization, poison rollback, and memory-level benchmarks. The extremely low research density of AS03 makes this a clear frontier.

\subsection{End-to-end VLAs need new observable security interfaces}
Hierarchical LLM planners expose explicit plans that can be checked. End-to-end VLAs may generate continuous actions directly, making the AS07--AS09 boundary implicit. Future architectures should deliberately expose security-relevant observables such as uncertainty, affordance justification, safety cost, action rationale, or structured latent certificates so that external monitors can validate behavior. If the system is an opaque vision-to-action black box, safety enforcement is pushed almost entirely to downstream control constraints and causal diagnosis becomes difficult.

\subsection{Multi-robot systems should assume Byzantine agents by default}
At fleet scale, one robot's output becomes another robot's input. Threat models should therefore include compromised robots, malicious messages, false claims, shared-memory poisoning, and colluding agents rather than assuming all collaborators are honest. Research is needed on Byzantine-resilient consensus, claim provenance, independent-sensing quorums, bounded trust delegation, compartmentalized shared memory, infection radius, and recovery time.

\subsection{Real-time constraints conflict with strong semantic verification}
Embodied defenses must run within the control cycle. Large VLM/LLM judges may reason semantically but are too slow or expensive for every high-frequency action. A promising design is a hierarchical monitor: fast geometric/control-barrier checks enforce hard constraints at low level, while foundation models perform lower-frequency semantic audits, with event-triggered escalation only when uncertainty or anomalies become high.

\subsection{Certified robustness must move from single frames to trajectories}
Traditional certified robustness often concerns one input and one prediction, while robot risk is determined by a trajectory. Future work should study trajectory-level certificates, reachable sets under adversarial observations, robust action chunks, closed-loop perturbation bounds, and cumulative risk. A perturbation that is tiny at every step but creates substantial long-horizon drift cannot be characterized adequately by a single-step $\ell_p$ norm.

\subsection{Privacy and security should be studied jointly}
This survey focuses primarily on integrity and physical safety, but VLALeaks demonstrates membership-inference risks in VLAs\cite{A042}. Real robots also continuously collect data about homes, workplaces, locations, voices, and people. Future work should address embodied privacy leakage, visual-memory privacy, cloud-inference exposure, multi-agent data sharing, and privacy-preserving telemetry, while examining whether privacy protection reduces the observability available to safety monitors.

\subsection{Build continuously updated open literature and evaluation infrastructure}
The field is growing too quickly for a static survey to remain complete for long. A more sustainable format combines a paper, open structured database, benchmark, and explicit taxonomy rules. Researchers should be able to submit new papers, update venues, correct multi-label coding, and add attack-chain annotations. A queryable corpus is more useful than a flat bibliography for tracking attack--defense gaps and evaluation bias over time.

\section{Discussion: Using the Taxonomy Without Overgeneralization}

\subsection{Model type is not an attack surface}
LLM, VLM, VLA, and WAM are architecture classes, not system entry points. A VLA may couple perception, reasoning, and action generation more tightly, but input, internal representation, and action authority can still be analyzed separately. Treating model type as the attack surface obscures distinct threat models that happen to use the same architecture.

\subsection{Final physical error is not a universal attack surface}
A robot ultimately taking a wrong action is a security consequence, not automatically an AS12 attack. Direct prompt injection should primarily be coded AS02, a visual patch AS05, and state spoofing AS06. AS12 becomes the primary entry only when trajectory, actuator command, or availability is directly manipulated. This principle is essential for avoiding double counting and for preserving causal interpretation.

\subsection{Security properties are orthogonal to attack surfaces}
The same surface can host multiple objectives. AS12 may contain trajectory-integrity attacks or freeze/DoS attacks; AS10 may leak credentials or modify ROS messages; AS01 may implant a backdoor or compromise model assets through a supply chain. Integrity, availability, privacy/confidentiality, and physical safety should therefore be coded independently of the entry surface.

\subsection{Paper count is not risk magnitude}
The abundance of AS05/AS09 papers partly reflects the popularity of VLA research and the relative ease of constructing quantitative visual/action benchmarks. The scarcity of AS03/AS10/AS11 papers does not imply low deployment risk. Persistent memory, cloud APIs, and fleet coordination may be highly important in production systems. Our statistics should be read as a map of academic attention, not a table of real-world attack probabilities.

\section{Conclusion}
Foundation models are transforming robots from closed control systems into agents that consume open information, maintain persistent state, reason autonomously, and obtain physical execution authority. Security can therefore no longer be partitioned cleanly into ``LLM jailbreak'', ``adversarial vision'', and ``robot network security''. The central questions are which representation an adversary can first control, which trust boundary that representation crosses, how malicious influence propagates into plans and actions, and whether an independent safety mechanism still exists before physical authority is granted.

Based on 58 attack records and 61 defense records through August 15, 2026, we introduce a five-layer, twelve-surface taxonomy using the first-compromised-trust-boundary principle and separate attack surface from mechanism, propagation path, security objective, and final effect. The resulting landscape shows that current attack research concentrates on multimodal perception and action interfaces, while defenses are even more concentrated downstream at action and execution boundaries. Context/long-term memory, middleware/cloud services, and multi-robot trust remain comparatively sparse. At the same time, world models, action chunks, persistent memory, and multi-agent coordination are introducing new security objects that conventional text-level guardrails cannot fully cover.

The central challenge for future embodied AI security is not merely to build a better jailbreak detector or a more robust visual encoder, but to establish a trusted architecture across provenance, state, reasoning, planning, action, and execution. Untrusted information should carry provenance; physical authority should be explicitly delegated; world state should be verifiable; planners and executors should be separated by privilege boundaries; runtime safety should balance latency with utility; and multi-robot systems should assume Byzantine participants. Only when attack research moves from model outputs to full causal chains, defenses move from point filters to compositional guarantees, and evaluation moves from textual ASR to closed-loop physical consequences will embodied AI security match the risks of real deployment.

\appendix
\section{Complete Coding Table of Attack Studies}
This appendix lists the 58 attack records in the current corpus. The table provides traceability between the taxonomy and the quantitative analysis in the main text. A single paper may carry multiple attack-surface codes.
\scriptsize
\setlength{\tabcolsep}{3pt}
\sloppy
\begin{longtable}{@{}p{0.55cm}p{5.15cm}p{0.8cm}p{2.1cm}p{2.0cm}p{1.65cm}p{2.45cm}@{}}
\caption{Complete attack-study coding table (through August 15, 2026).}\\
\toprule
No. & Paper / Method & Year & Attack mechanism & Attack surfaces & Model & Validation environment\\
\midrule
\endfirsthead
\toprule
No. & Paper / Method & Year & Attack mechanism & Attack surfaces & Model & Validation environment\\
\midrule
\endhead
1 & On the Vulnerability of LLM/VLM-Controlled Robotics & 2024 & Adversarial input / robustness attack & AS02;AS05;AS08 & LLM/VLM & Simulation / partial real-world tests \\
2 & Exploring the Robustness of Decision-Level Through Adversarial Attacks on LLM-Based Embodied Models (EIRAD) & 2024 & Adversarial prompt / targeted and untargeted attacks & AS02;AS08 & LLM/VLM & Simulation \\
3 & Can We Trust Embodied Agents? Exploring Backdoor Attacks against Embodied LLM-based Decision-Making Systems (BALD) & 2025 & Backdoor: word/scenario/knowledge injection & AS01;AS02;AS06 & LLM & Simulation \\
4 & BadRobot: Jailbreaking Embodied LLMs in the Physical World & 2024 & Jailbreak / language--action misalignment & AS02;AS07;AS08;AS09 & LLM & Simulation + real robot \\
5 & A Study on Prompt Injection Attack Against LLM-Integrated Mobile Robotic Systems & 2024 & Prompt Injection & AS02;AS08 & LLM & Simulation / mobile robot \\
6 & Compromising LLM Driven Embodied Agents with Contextual Backdoor Attacks & 2025 & ICL/Contextual Backdoor & AS03;AS01 & LLM/VLM & Simulation / partial real-world tests \\
7 & Jailbreaking LLM-Controlled Robots (RoboPAIR) & 2025 & Automated jailbreak / adversarial prompt search & AS02;AS08;AS09 & LLM & Simulation + real robot \\
8 & TrojanRobot / Robot Collapse: Supply-Chain Backdoor Attacks Against VLM-based Robotic Manipulation & 2025 & Perception-model backdoor / supply-chain attack & AS01;AS05 & VLM & Real robot \\
9 & POEX: Understanding and Mitigating Policy Executable Jailbreak Attacks against Embodied AI & 2024 & Policy-Executable Jailbreak & AS02;AS08;AS09 & LLM & Simulation \\
10 & BEAT: Visual Backdoor Attacks on MLLM Embodied Decision Making via Contrastive Trigger Learning & 2025 & Visual backdoor / contrastive trigger learning & AS01;AS05;AS08 & VLM/MLLM & Simulation \\
11 & BadNAVer: Jailbreaking Vision-Language Navigation Agents & 2025 & Navigation jailbreak & AS02;AS08 & VLM & Simulation \\
12 & Adversarial Attacks on Robotic Vision Language Action Models & 2025 & Language adversarial attack / action-space hijacking & AS02;AS07;AS09 & VLA & Simulation + real robot \\
13 & BadVLA: Towards Backdoor Attacks on Vision-Language-Action Models via Objective-Decoupled Optimization & 2025 & VLA backdoor / objective-decoupled optimization & AS01;AS05;AS09 & VLA & Simulation + real robot \\
14 & Human/Model-Agnostic Adversarial Attacks and Defenses for Vision-Language-Action Models & 2025 & Model-agnostic adversarial attack & AS05;AS09 & VLA & Simulation \\
15 & Goal-Oriented Backdoor Attack Against Vision-Language-Action Models via Physical Objects (GoBA) & 2025 & Physical-object backdoor & AS01;AS04;AS05;AS09 & VLA & Simulation + real robot \\
16 & DropVLA: An Action-Level Backdoor Attack on Vision-Language-Action Models & 2026 & Action-level Backdoor & AS01;AS09 & VLA & Simulation \\
17 & Clean-Action Backdoor Attacks on Vision-Language-Action Models via Sequential Error Exploitation & 2025 & Clean-label/Sequential Backdoor & AS01;AS09 & VLA & Simulation \\
18 & AttackVLA: Benchmarking Adversarial and Backdoor Attacks on Vision-Language-Action Models & 2025 & Unified adversarial/backdoor attacks; long-horizon target actions & AS01;AS02;AS05;AS09 & VLA & Simulation \\
19 & When Alignment Fails: Multimodal Adversarial Attacks on Vision-Language-Action Models (VLA-Fool) & 2025 & Cross-modal adversarial attack & AS05;AS07;AS09 & VLA & Simulation \\
20 & UPA-RFAS: Universal Physical Adversarial Patch Attacks on Vision-Language-Action Models & 2025 & Universal transferable physical patch & AS05;AS09 & VLA & Simulation + real robot \\
21 & FreezeVLA: Action-Freezing Attacks Against Vision-Language-Action Models & 2025 & Visual adversarial / availability attack & AS05;AS12 & VLA & Simulation \\
22 & The Shawshank Redemption of Embodied AI: Understanding and Benchmarking Indirect Environmental Jailbreaks & 2025 & Indirect Environmental Jailbreak & AS04;AS05;AS08 & VLM/Embodied Agent & Simulation \\
23 & SABER: A Stealthy Agentic Black-Box Attack Framework for Vision-Language-Action Models & 2026 & Agentic Black-box Attack & AS02;AS05;AS09 & VLA & Simulation / physically transferable \\
24 & CHAI: Command Hijacking Against Embodied AI & 2025 & Physical visual prompt injection & AS04;AS05;AS08 & VLM/Embodied Agent & Simulation / physical scenes \\
25 & Manipulating Multimodal Agents via Cross-Modal Prompt Injection (CrossInject) & 2025 & Cross-modal Prompt Injection & AS03;AS04;AS05;AS08 & VLM/MLLM & Simulation \\
26 & Extended to Reality: Prompt Injection in 3D Environments (PI3D) & 2026 & 3D-environment prompt injection & AS04;AS05 & VLM/Agent & Simulation / 3D environment \\
27 & State Backdoor: Towards Stealthy Real-world Poisoning Attack on Vision-Language-Action Model in State Space & 2026 & State-space Backdoor & AS01;AS06;AS09 & VLA & Simulation + real robot \\
28 & SilentDrift: Exploiting Action Chunking for Stealthy Backdoor Attacks on Vision-Language-Action Models & 2026 & Action-chunk Backdoor & AS01;AS09;AS12 & VLA & Simulation \\
29 & Jailbreaking Embodied LLMs via Action-level Manipulation (Blindfold) & 2026 & Adversarial Proxy Planning / Action Manipulation & AS07;AS08;AS09 & LLM & Simulation + real robot \\
30 & Altered Thoughts, Altered Actions: Probing Chain-of-Thought Vulnerabilities in VLA Robotic Manipulation & 2026 & CoT/Reasoning Manipulation & AS07;AS08;AS09 & VLA & Simulation \\
31 & TRAP: Hijacking VLA CoT-Reasoning via Adversarial Patches & 2026 & Physical patch to CoT hijacking & AS05;AS07;AS09 & VLA & Simulation + physical tests \\
32 & Tex3D: Physical 3D Texture Attacks on Vision-Language-Action Models & 2026 & 3D Texture/Physical Adversarial Attack & AS05;AS09 & VLA & Simulation / physically realizable \\
33 & From Prompt to Physical Action: Structured Backdoor Attacks on LLM-Mediated Robotic Control Systems & 2026 & Structured-control backdoor / LoRA & AS01;AS09;AS10 & LLM & Simulation + real robot \\
34 & FlowHijack: Backdoor Attacks on Flow-Matching Vision-Language-Action Models & 2026 & Action-generation Dynamics Backdoor & AS01;AS07;AS09 & VLA & Simulation \\
35 & Semantic Denial of Service in LLM-Controlled Robots & 2026 & Semantic DoS / Safety-trigger Abuse & AS02;AS04;AS12 & LLM & Simulation + real robot/audio \\
36 & JailWAM: Jailbreaking World Action Models in Robot Control & 2026 & World-Action Model Jailbreak & AS02;AS07;AS09 & WAM/World-Action Model & Simulation \\
37 & PINA: Prompt Injection Attack against Navigation Agents & 2026 & Adaptive Prompt Injection & AS02;AS08 & LLM/VLM Agent & Simulation \\
38 & RIPA: Sensory-Vector Prompt Injection Attacks on LLM-Controlled ROS 2 Robots & 2026 & Vision/OCR, audio/STT, and LiDAR-state injection & AS04;AS05;AS06;AS10 & LLM & ROS2 experimental system \\
39 & Phantom Menace: Exploring and Enhancing the Robustness of VLA Models Against Physical Sensor Attacks & 2026 & Physical sensor-signal injection & AS05;AS12 & VLA & Simulation + real robot \\
40 & Partially Observable Adversarial Patch Attacks on Vision-Language-Action Models & 2026 & Partially observable physical patch & AS05;AS09 & VLA & Simulation / physical tests \\
41 & When World Models Dream Wrong: Physical-Conditioned Adversarial Attacks Against World Models & 2026 & Physically conditioned adversarial attack & AS05;AS06;AS07 & World Model & Simulation \\
42 & VLALeaks: Membership Inference Attacks against Vision-Language-Action Models & 2026 & Membership Inference & AS01 & VLA & Offline model evaluation \\
43 & When Prompts Control Robots: Prompt Injection Attacks in Multi-Agent Robotic Systems & 2026 & Direct/Indirect Prompt Injection + Cross-Agent Propagation & AS02;AS04;AS11 & LLM & Multi-agent simulation \\
44 & Propagating Unsafe Actions in LLM-Controlled Multi-Robot Collaboration via Single Robot Compromise (InfectBot) & 2026 & Single-robot compromise to multi-robot propagation & AS11 & LLM & Multi-robot simulation \\
45 & When Coordination Becomes a Threat: Communication Attacks in LLM-Controlled Multi-Robot Systems & 2026 & External Entry / Privileged In-System Communication Attack & AS11 & LLM & Multi-robot simulation \\
46 & Hijacking Robots with a Piece of Paper: A Systematic Study of Physical Prompt Injection in VLM-Controlled Robots & 2026 & Physical Prompt Injection & AS04;AS05;AS08 & VLM & Real robot \\
47 & VLAGuard / VASA: Visuomotor Attention-guided Semantic Attack & 2026 & Attention-Hijacking Physical Patch & AS05;AS07;AS09 & VLA & Simulation + 2,000 real-world trials \\
48 & DRIFT: Derailing Trajectories of Flow-Matching VLAs with Adversarial Patches & 2026 & Universal physical patch / trajectory derailment & AS05;AS09 & VLA & Simulation + physical tests \\
49 & Structure-Aware Robust Fine-Tuning / AGSD Attack & 2026 & Attention-Guided Spatial Distraction (AGSD) & AS05;AS07;AS09 & VLA & Simulation + physical tests \\
50 & Hidden in Plain Sight: Diffusion-Based Unrestricted Robotic Attacks on Vision-Language-Action Models (DURA) & 2026 & Diffusion-based Natural Adversarial Patch & AS05;AS09 & VLA & Simulation + real robot \\
51 & From Prompts to Motors: Man-in-the-Middle Risks for LLM-Connected Robotic Systems & 2025 & MITM / API Message Manipulation & AS10;AS09 & LLM & System experiments \\
52 & Net-GPT: a LLM-Empowered Man-in-the-Middle Chatbot for Unmanned Aerial Vehicle & 2023 & LLM-assisted MITM & AS10 & LLM & System experiments \\
53 & Supply Chain Exploitation of Secure ROS 2 Systems: A Proof-of-Concept on Autonomous Platform Compromise via Keystore Exfiltration & 2025 & ROS2 Supply-Chain / Credential Theft & AS10;AS01 & - & Real robot / system \\
54 & On the (In)Security of Secure ROS2 & 2022 & ROS2/DDS Security Bypass & AS10 & - & System experiments \\
55 & Automated Discovery of Semantic Attacks in Multi-Robot Navigation Systems & 2025 & Semantic/Systematic Multi-Robot Attack Discovery & AS11;AS12 & - & Simulation / system \\
56 & PhysPatch: A Physically Realizable and Transferable Adversarial Patch Attack for Multimodal Large Language Models-based Autonomous Driving Systems & 2026 & Physically realizable / transferable adversarial patch & AS05;AS07;AS08 & MLLM/VLM & Physically realizable \\
57 & Inject Once, Survive Later: Backdooring Vision-Language-Action Models to Persist Through Downstream Fine-tuning (INFUSE) & 2026 & Persistent backdoor / fine-tuning-insensitive module injection & AS01;AS05;AS09 & VLA & Simulation + real robot \\
58 & If you're waiting for a sign... that might not be it! Trust Boundary Confusion from Visual Injections on Vision-Language Agentic Systems & 2026 & Visual Injection / Trust-Boundary Confusion & AS04;AS05;AS08 & VLM/LVLM & Simulation / embodied scenarios \\
\bottomrule
\end{longtable}
\normalsize

\section{Complete Coding Table of Defense Studies}
\scriptsize
\setlength{\tabcolsep}{3pt}
\sloppy
\begin{longtable}{@{}p{0.55cm}p{5.1cm}p{0.8cm}p{2.4cm}p{2.0cm}p{1.7cm}p{2.25cm}@{}}
\caption{Complete defense-study coding table (through August 15, 2026).}\\
\toprule
No. & Paper / Method & Year & Defense mechanism & Covered surfaces & Model & Validation environment\\
\midrule
\endfirsthead
\toprule
No. & Paper / Method & Year & Defense mechanism & Covered surfaces & Model & Validation environment\\
\midrule
\endhead
1 & Plug in the Safety Chip: Enforcing Constraints for LLM-driven Robot Agents & 2024 & Formal constraints / LTL safety chip & AS02;AS08;AS09 & LLM & Simulation + real robot \\
2 & Ensuring Safety in LLM-Driven Robotics: A Cross-Layer Sequence Supervision Mechanism & 2024 & Cross-layer sequence supervision & AS02;AS08;AS09 & LLM & Simulation / robot \\
3 & SafeEmbodAI: Securing LLM-Controlled Mobile Robots against Prompt Injection & 2024 & Secure Prompting + State Management + Safety Validation & AS02;AS03;AS08 & LLM & Simulation / mobile robot \\
4 & SELP: Generating Safe and Efficient Task Plans for Robot Agents with Large Language Models & 2024 & Safety-aware decoding / unsafe-action pruning & AS08;AS09 & LLM & Simulation \\
5 & Safe LLM-Controlled Robots with Formal Guarantees via Reachability Analysis & 2025 & Reachability / Formal Guarantee & AS08;AS09;AS12 & LLM & Simulation \\
6 & SafePlan: Leveraging Formal Logic and Chain-of-Thought Reasoning for Enhanced Safety in LLM-based Robotic Task Planning & 2025 & Prompt Sanity + Invariant + Preconditions/Postconditions & AS02;AS08;AS09 & LLM & Simulation \\
7 & Safe-BeAl / Safe-Align: A Framework for Benchmarking and Aligning Task Planning of LLM-based Embodied Agents & 2025 & Safety Benchmark + Alignment & AS02;AS08 & LLM & Simulation \\
8 & Safety Guardrails for LLM-Enabled Robots (RoboGuard) & 2026 & Contextual safety rules to temporal logic to safe-control synthesis & AS06;AS08;AS09 & LLM & Simulation + real robot \\
9 & Concept Enhancement Engineering (CEE) for Safeguarding Embodied LLMs & 2025 & Hidden-state Safety Steering & AS07;AS08 & LLM & Simulation \\
10 & J-DAPT: Preventing Robotic Jailbreaking via Multimodal Domain Adaptation & 2025 & Multimodal jailbreak detector + domain adaptation & AS02;AS04;AS05 & LLM/VLM & Multi-scenario simulation \\
11 & SafeMind: Benchmarking and Mitigating Safety Risks in Embodied LLM Agents & 2025 & Cross-stage safety agent / factual--causal--temporal constraints & AS04;AS05;AS06;AS08;AS09 & LLM/VLM & Simulation \\
12 & RoboSafe: Safeguarding Embodied Agents via Executable Safety Logic & 2025 & Executable Safety Logic + Memory + Predictive/Reflective Reasoning & AS06;AS08;AS09;AS12 & LLM/VLA & Simulation + real robot \\
13 & PROTEA: Securing Robot Task Planning and Execution & 2026 & Independent LLM-as-a-Judge / Plan Verification & AS08;AS09 & LLM & Simulation \\
14 & Pre-Execution Safety Gate \& Task Safety Contracts for LLM-Based Robot Task Planning (SafeGate) & 2026 & Safety Gate + Contract + SMT/Z3 & AS02;AS06;AS08;AS09 & LLM & AI2-THOR + real robot \\
15 & LogicGuard: Improving Embodied LLM Agents Through Temporal Logic-Based Critics & 2026 & Temporal-Logic Critic & AS08;AS09 & LLM & Simulation \\
16 & SAFE: Multitask Failure Detection for Vision-Language-Action Models & 2025 & Latent-feature Failure Detection & AS07;AS09;AS12 & VLA & Simulation \\
17 & AHA: A Vision-Language-Model for Detecting and Reasoning over Failures in Robotic Manipulation & 2024 & VLM Failure Detection + Reasoning & AS09;AS12 & VLM & Real robot / simulation \\
18 & Guardian: Detecting Robotic Planning and Execution Errors with Vision-Language Models & 2025 & External VLM Monitor & AS08;AS09;AS12 & VLM & Simulation / real robot \\
19 & Vision-Language Models for Robot Success Detection & 2024 & VLM Success/Failure Detection & AS09;AS12 & VLM & Real robot / simulation \\
20 & FPC-VLA: A Vision-Language-Action Framework with a Supervisor for Failure Prediction and Correction & 2026 & Supervisor + Failure Prediction/Correction & AS09;AS12 & VLA & Simulation / robot \\
21 & RoVer: Runtime Verification for Vision-Language-Action Policies & 2025 & Runtime Verification/Monitor & AS08;AS09;AS12 & VLA & Simulation \\
22 & Ask Before You Act: Token-Level Uncertainty for Intervention in Vision-Language-Action Models & 2025 & Token-level Uncertainty + Human Intervention & AS07;AS09 & VLA & Simulation \\
23 & SafeDec: Constrained Decoding for Safe Autoregressive Embodied Policies & 2025 & Constrained Decoding & AS09;AS12 & VLA/Autoregressive Policy & Simulation \\
24 & Run-time Observation Interventions Make Vision-Language-Action Models More Visually Robust (BYOVLA) & 2025 & Observation Masking/Intervention & AS05;AS09 & VLA & Simulation + real robot \\
25 & Safe-VLN: Collision Avoidance for Vision-and-Language Navigation of Autonomous Robots Operating in Continuous Environments & 2024 & Occupancy-aware Collision Avoidance & AS06;AS08;AS12 & VLM/VLN & Simulation / robot \\
26 & Affordance Field Intervention: Enabling VLAs to Escape Memory Traps in Robotic Manipulation & 2025 & Affordance-field rollback/recovery & AS06;AS09;AS12 & VLA & Simulation \\
27 & REFLECT: Summarizing Robot Experiences for Failure Explanation and Correction & 2023 & Experience Summary + Failure Reflection & AS08;AS09;AS12 & LLM/VLM & Real robot / simulation \\
28 & FailSafe: Reasoning and Recovery from Failures in Vision-Language-Action Models & 2025 & External VLM Reasoning + Replanning & AS08;AS09;AS12 & VLA/VLM & Simulation + real robot \\
29 & Causal Scene Narration with Runtime Safety Supervision for Vision-Language-Action Driving & 2026 & Causal Scene Narration / Intent-Constraint Monitor & AS05;AS06;AS08 & VLA/VLM & Simulation \\
30 & VLSA: Vision-Language-Action Models with Plug-and-Play Safety Constraint Layer & 2025 & Plug-and-play Safety Constraint Layer & AS09;AS12 & VLA & Simulation \\
31 & SafeVLA: Towards Safety Alignment of Vision-Language-Action Model via Constrained Learning & 2025 & CMDP/Constrained Safe RL & AS01;AS09 & VLA & Simulation \\
32 & Safety Optimized Reinforcement Learning via Multi-Objective Policy Optimization (SORL) & 2024 & Safety Critic + Multi-objective Optimization & AS01;AS09 & Policy/VLA-adjacent & Simulation \\
33 & Human-Assisted Robotic Policy Refinement via Action Preference Optimization (APO) & 2025 & Human Intervention  to  Action Preference & AS09 & VLA/Robot Policy & Robot experiments \\
34 & Human-in-the-loop Online Rejection Sampling for Robotic Manipulation (Hi-ORS) & 2025 & Human Correction + Rejection Sampling & AS09 & VLA/Robot Policy & Robot experiments \\
35 & EvoVLA: Self-Evolving Vision-Language-Action Model & 2025 & Self-evolving Policy Improvement & AS01;AS09 & VLA & Simulation / robot \\
36 & Generative Scenario Rollouts for End-to-End Autonomous Driving (GSR) & 2026 & Generative Safety Scenario Rollout & AS01;AS05;AS09 & VLA/Driving & Simulation \\
37 & Pedagogical Alignment for Vision-Language-Action Models & 2026 & Pedagogical Safety Alignment & AS01;AS09 & VLA & Simulation \\
38 & Safe-Night VLA: Seeing the Unseen via Thermal-Perceptive Vision-Language-Action Models for Safety-Critical Manipulation & 2026 & Thermal Perception + CBF-QP Safety Filter & AS05;AS09;AS12 & VLA & Real robot / simulation \\
39 & VLA-Forget: Vision-Language-Action Unlearning for Embodied Foundation Models & 2026 & Safety Unlearning & AS01;AS09 & VLA & Simulation \\
40 & RobustVLA: Robustness-Aware Reinforcement Post-Training for Vision-Language-Action Models & 2025 & Robustness-aware RL Post-training & AS05;AS09 & VLA & Simulation \\
41 & RETAIN: Robust Finetuning of Vision-Language-Action Robot Policies via Parameter Merging & 2025 & Robust Fine-tuning + Parameter Merging & AS01;AS05;AS09 & VLA & Simulation \\
42 & MergeVLA: Robust Vision-Language-Action Models via Model Merging & 2025 & Model Merging / Robustness Composition & AS01;AS05;AS09 & VLA & Simulation \\
43 & Horcrux: Robustness/Recovery for Vision-Language-Action Policies & 2025 & Redundant/Recovery Robustness Mechanism & AS05;AS09;AS12 & VLA & Simulation \\
44 & SAFE-Dict: Concept-Based Dictionary Learning for Inference-Time Safety in Vision Language Action Models & 2026 & Concept Dictionary / Representation Intervention & AS07;AS09 & VLA & Simulation \\
45 & TrustVLA: Mechanism-Guided Inference-Time Defense Against Vision-Language-Action Backdoors & 2026 & Causal-footprint Detection + Trigger Localization/Inpainting & AS01;AS05;AS07;AS09 & VLA & Simulation \\
46 & VLAGuard / APFT: Attention-Protective Fine-Tuning & 2026 & Attention-Protective Fine-Tuning & AS05;AS07;AS09 & VLA & Simulation + 2,000 real-world trials \\
47 & Structure-Aware Robust Fine-Tuning (SARF) & 2026 & Structure-aware Robust Fine-Tuning & AS05;AS07;AS09 & VLA & Simulation + physical tests \\
48 & ChromaGuard: Defending VLA Models against Color/Illumination-Induced Blindness & 2026 & Color-aware Robustness Guard & AS05;AS07;AS09 & VLA & Simulation / physical tests \\
49 & Phantom Menace Robustness Enhancement & 2026 & Sensor-Attack Robust Training & AS05;AS12 & VLA & Simulation + real robot \\
50 & RIPA Hybrid Semantic Firewall & 2026 & Rule + Semantic Firewall & AS02;AS04;AS05;AS06 & LLM & ROS2 experimental system \\
51 & Physical Prompt Injection Mitigations: Prompt Defense, Two-Stage Verification, and Text Masking & 2026 & Prompt Hardening + Verification + Text Masking & AS04;AS05;AS08 & VLM & Real robot \\
52 & Per-Agent LLM Isolation for Multi-Agent Robotic Prompt Injection & 2026 & Per-Agent Context/LLM Isolation & AS11;AS03 & LLM & Multi-agent simulation \\
53 & Claim Provenance and Verification (CPV) Gate & 2026 & Provenance + Independent Claim Verification & AS11 & LLM & Multi-robot simulation \\
54 & Structured Backdoor Verification / Output-Schema Defense & 2026 & Structured Output Verification & AS01;AS09;AS10 & LLM & Simulation + real robot \\
55 & RoboRebound: Multi-Robot System Defense with Bounded-Time Interaction & 2025 & Bounded-time Multi-Robot Resilience & AS11;AS12 & - & Multi-robot system \\
56 & Byzantine Resilience at Swarm Scale: A Decentralized Blocklist Protocol from Inter-Robot Accusations & 2023 & Decentralized Trust/Blocklist & AS11 & - & swarm simulation \\
57 & Trusted Operations of a Military Ground Robot in the Face of Man-in-the-Middle Cyberattacks Using Deep Learning CNNs & 2023 & MITM Detection & AS10 & - & Real robot / network experiments \\
58 & When Attention Betrays: Erasing Backdoor Attacks in Robotic Policies by Reconstructing Visual Tokens (Bera) & 2026 & Abnormal-attention detection + trigger localization/reconstruction & AS01;AS05;AS07;AS09 & VLA & Simulation / multiple embodied platforms \\
59 & Trust-Boundary-Aware Multi-Agent Defense for Visual Injections & 2026 & Perception--decision decoupling + multi-agent trust evaluation & AS04;AS05;AS08 & VLM/LVLM & Simulation / embodied scenarios \\
60 & Model-Agnostic Adversarial Defense for Vision-Language-Action Models & 2025 & Model-agnostic adversarial defense & AS05;AS09 & VLA & Simulation \\
61 & Towards Safe Robot Foundation Models Using Inductive Biases & 2025 & Safety-oriented Inductive Biases & AS01;AS07;AS09 & Robot Foundation Model/VLA & Simulation \\
\bottomrule
\end{longtable}
\normalsize

\section{Quick Decision Rules for Overlapping Attack Scenarios}
For a composite attack that is difficult to classify, answer the following questions in order and use the first satisfied \emph{direct attack point} as the primary entry surface:
\begin{enumerate}
  \item Was malicious influence written into training, fine-tuning, or model/component distribution? If yes, AS01.
  \item During deployment, was the malicious information directly supplied through the currently authorized user-task channel? If yes, AS02.
  \item Did it originate from ICL examples, memory, RAG, interaction history, or tool returns? If yes, AS03.
  \item Is it semantic content that physically exists in the scene and is correctly read by the sensor? If yes, AS04.
  \item Does the adversary directly alter image, audio, optical, LiDAR, or another raw sensor signal? If yes, AS05.
  \item Does the adversary directly alter world state, localization, object relations, affordances, joint state, or feedback? If yes, AS06.
  \item Does the adversary directly target CoT, attention, hidden representations, or cross-modal alignment? If yes, AS07.
  \item Does the adversary directly modify the goal, task decomposition, plan, path, or high-level policy? If yes, AS08.
  \item Does the adversary directly modify code, JSON, API/tool calls, skills, or action tokens/chunks? If yes, AS09.
  \item Does the attack exploit ROS/DDS, networking, cloud services, credentials, or a software/service chain? If yes, AS10.
  \item Does the malicious information propagate directly from another robot/agent or a shared collaboration state? If yes, AS11.
  \item Does the adversary directly act on trajectories, actuators, halt/freeze mechanisms, or runtime availability? If yes, AS12.
\end{enumerate}

\end{document}